\documentclass[10pt,journal,compsoc]{IEEEtran}

\ifCLASSOPTIONcompsoc
  \usepackage{cite}
\else
  \usepackage{cite}
\fi

\ifCLASSINFOpdf
\else
\fi

\usepackage{graphicx}
\usepackage{comment}
\usepackage{amsmath,amssymb} 
\allowdisplaybreaks[1]
\usepackage{color}
\usepackage{booktabs}
\usepackage{array}
\usepackage[table]{xcolor}
\usepackage[caption=false,font=footnotesize]{subfig}
\usepackage{diagbox}
\usepackage{multirow}
\usepackage{hyperref}
\usepackage{relsize}

\newcommand{\eqnref}[1]{Equation~(\ref{eqn:#1})}
\newcommand{\figref}[1]{Fig.~\ref{fig:#1}}

\newcommand{\secref}[1]{Section~\ref{sec:#1}}
\newcommand{\appref}[1]{Appendix~\ref{app:#1}}

\newcommand\numberthis{\addtocounter{equation}{1}\tag{\theequation}}

\newcommand\independent{\protect\mathpalette{\protect\independenT}{\perp}}
\def\independenT#1#2{\mathrel{\rlap{$#1#2$}\mkern2mu{#1#2}}}

\definecolor{cobalt}{rgb}{0.0, 0.28, 0.67}
\hypersetup{
colorlinks=true,
urlcolor=black,
linkcolor=cobalt,
citecolor=cobalt}

\newcommand{\graphrnn}{
\begin{figure*}[t]
    \centering
    \includegraphics[trim={0in 0in 0in 0in},width=1\textwidth]{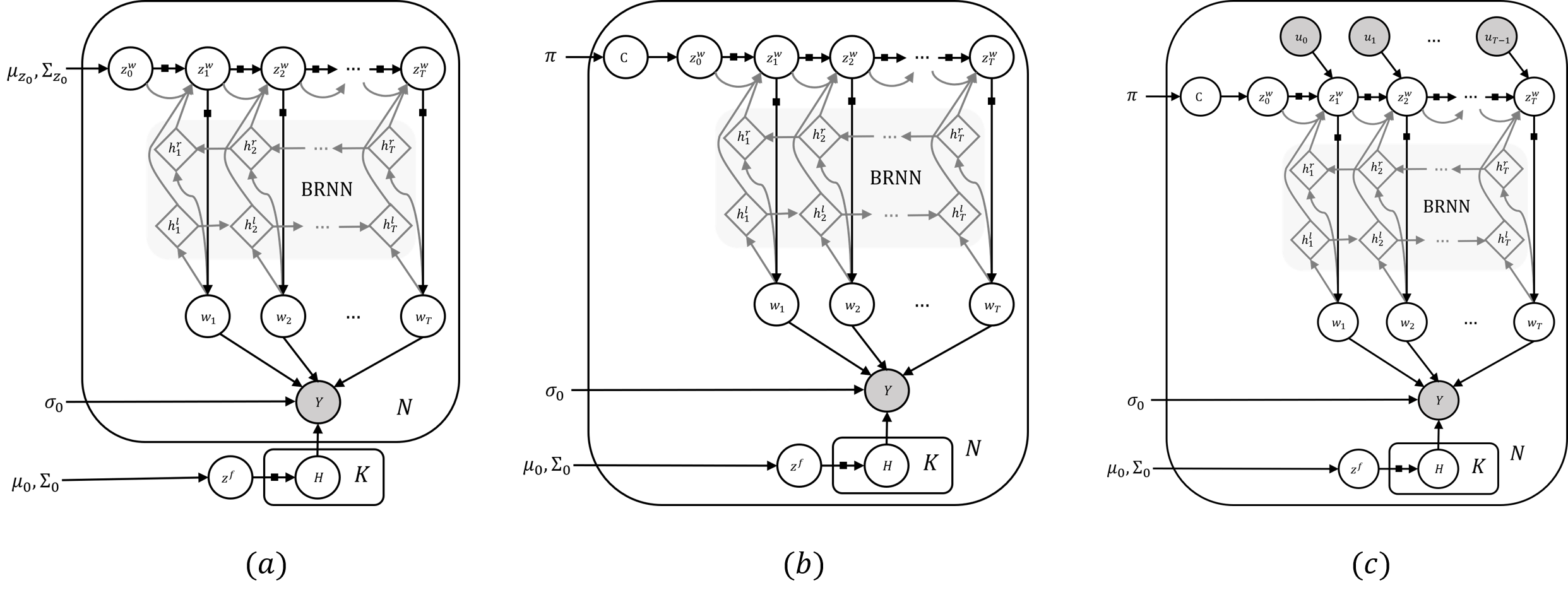}
    \caption{Graphical model representation for the three variants of DMSTF. All three variants incorporate a deep generative Markovian prior, $p_\theta(z^w_t|z^w_{t-1})$, to represent temporal variations in $W$. The $K$ spatial factors are conditioned on a shared latent, $z^{f}$. Latent nodes and observations are represented by solid and gray-shaded circles, respectively. The solid black squares denote nonlinear mappings parameterized by neural networks. Gray lines represent variational distribution. \textbf{(a)} This variant assumes that the factor parameters $H_n$ and latents $z_n^f$ do not vary across instances, but are instead shared at the corpus-level. \textbf{(b)} This variant introduces an additional discrete latent $c$, which encourages a multimodal distribution for the temporal generative model, and serves to cluster instances. \textbf{(c)} This variant introduces a sequence of observed control variables, $u_{0:T-1}$, that govern the temporal distribution as $p_\theta(z^w_t|z^w_{t-1}, u_{t-1})$.}
    \label{fig:rnn}
\end{figure*}
}

\newcommand{\synres}{
\begin{figure*}[t]
    \centering
    \includegraphics[trim={0in 0in 0in 0in},width=1\textwidth]{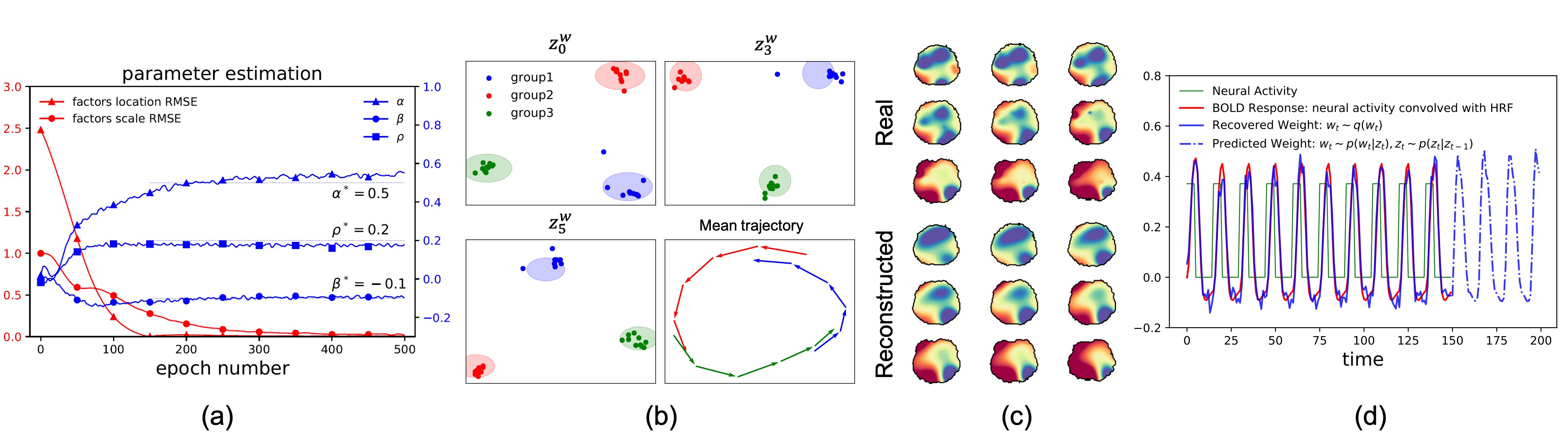}
    \caption{\textbf{(a)} DMSTF recovered the actual parameters of a nonlinear dynamical system in our toy example. \textbf{(b)} DMSTF recovered the three clusters of activation in our synthetic fMRI dataset, unsupervised. The mean dynamical trajectory (composed of three consecutive rotational dynamics) shows the inferred trajectory of each cluster mean over time in the temporal latent, and is consistent with the periodic activation of sources in data clusters. \textbf{(c)} Real and reconstructed brain images. \textbf{(d)} The learned generative model's predictions for a selected activation source show that DMSTF encoded the nonlinear hemodynamic response function in its deep temporal generative model.}
    \label{fig:synres}
\end{figure*}
}

\newcommand{\autres}{
\begin{figure*}[t]
    \centering
    \includegraphics[trim={0in 0in 0in 0in},width=\textwidth]{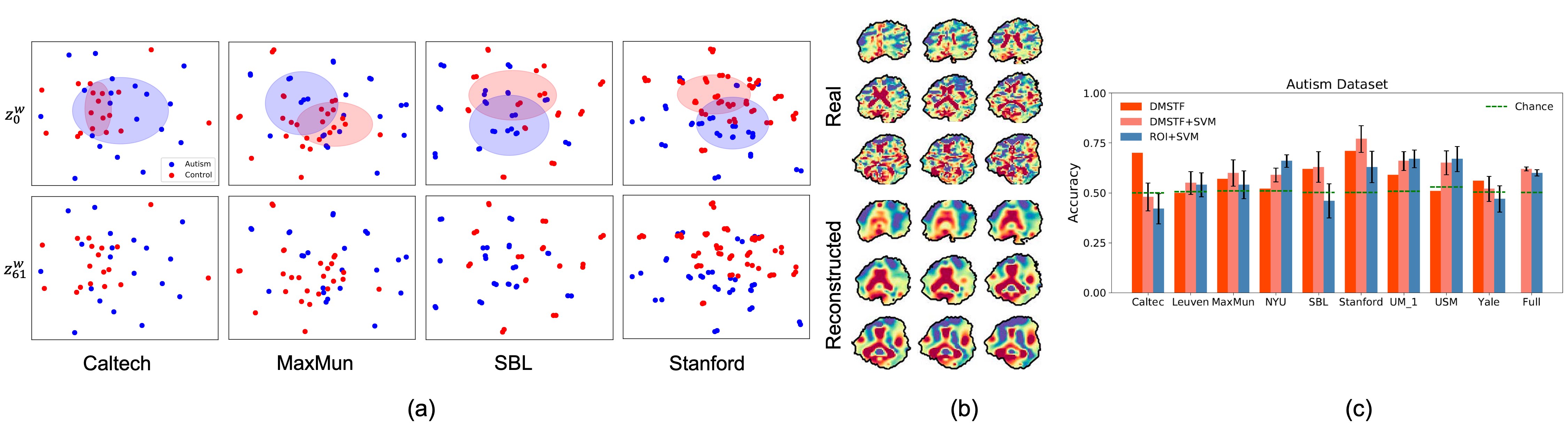}
    \caption{\textbf{(a)} DMSTF's clustering results show that the ASD and control groups can be partially separated. \textbf{(b)} Real and reconstructed brain images, showing the smoothing given by sparse factorization. \textbf{(c)} A downstream classification task showed that \texttt{DMSTF} and \texttt{DMSTF+SVM} outperformed regions of interest \texttt{(ROI)+SVM} in the Caltech, MaxMun, SBL, Stanford, and Yale subsets of the data. \texttt{ROI+SVM} performed better in the NYU subset.}
    \label{fig:autres}
\end{figure*}
}

\newcommand{\depres}{
\begin{figure*}[t]
    \centering
    \includegraphics[trim={0in 0in 0in 0in},width=\textwidth]{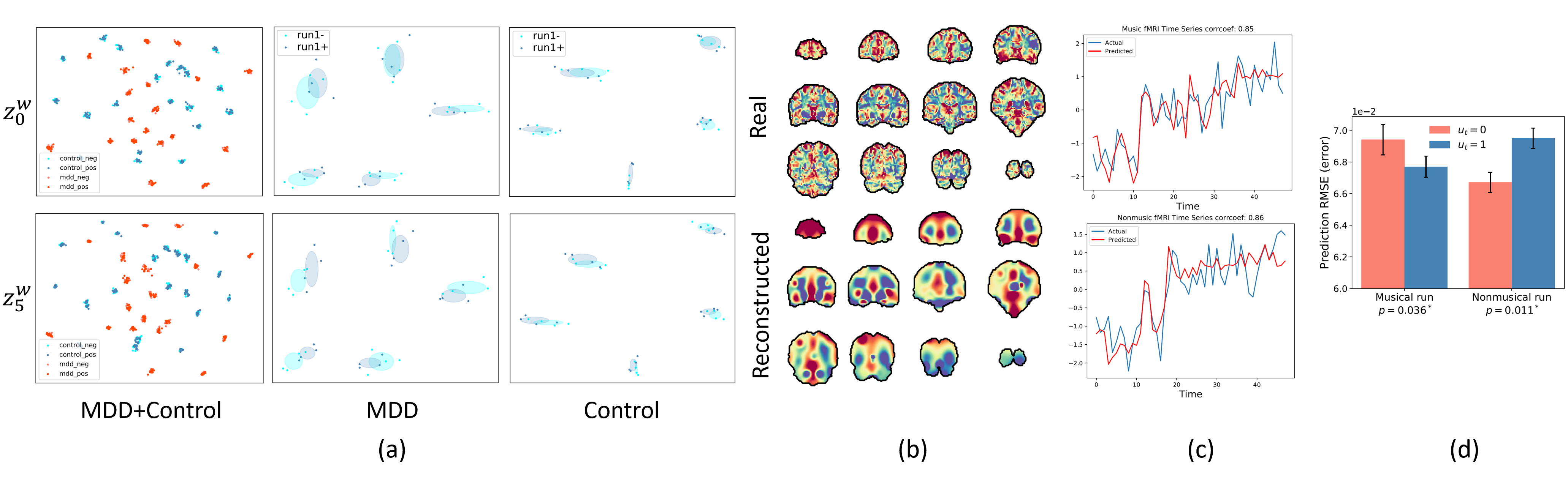}
    \caption{\textbf{(a), Left}: Training DMSTF clustered together temporal latent variables associated with each subject without supervision, while partially separating clusters of points associated with the MDD group from those associated with the control group. The MDD group appears more concentrated into the center of the temporal latent space, while the control group have their temporal latent variables dispersed more broadly across the latent space. \textbf{(a), Middle, Right}: DMSTF enabled us to partially separate ``positive'' and ``negative'' stimuli per-subject with Gaussian clusters. \textbf{(b)} Real and reconstructed brain images. \textbf{(c, d)} The control variable $u_t$ is a good predictor of temporal sequences in the trained model, with $u_t=0$ fitting nonmusical sequences and $u_t=1$ fitting musical sequences.  Example fMRI time-series from both musical and nonmusical trials are shown in \textbf{(c)}.}
    \label{fig:depres}
    \vspace{-.3cm}
\end{figure*}
}

\newcommand{\trares}{
\begin{figure*}[t]
    \centering
    \includegraphics[trim={0in 0in 0in 0in},width=1\textwidth]{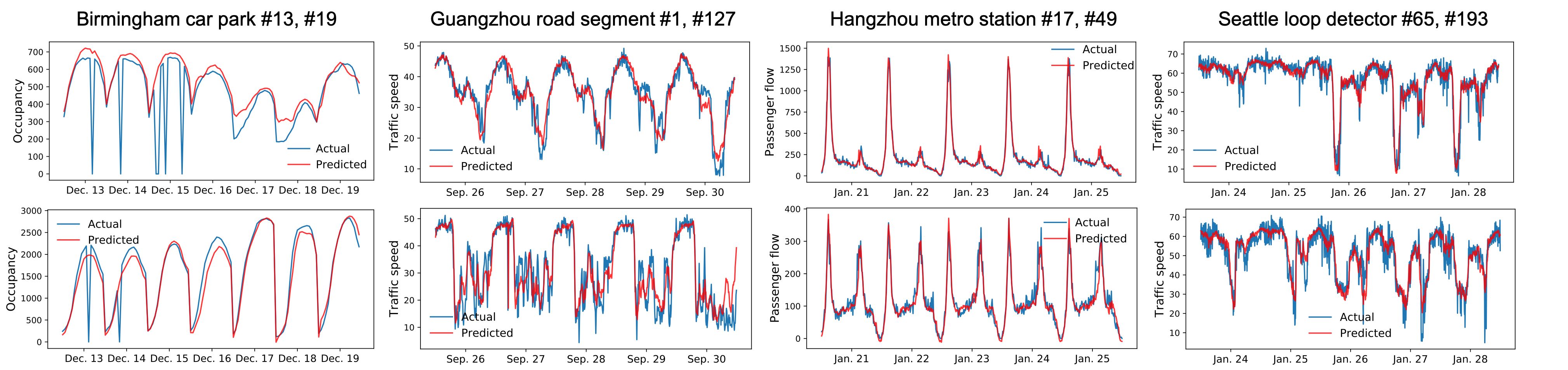}
    \caption{Predicted time-series for two sample locations in the test-set of each traffic data. Note that the Birmingham and Guangzhou datasets are missing some values, which prediction fills in.}
    \label{fig:trares}
\end{figure*}
}

\newcommand{\picaut}{
\begin{figure}[!htp]
    \centering
    \includegraphics[trim={0in 0in 0in 0in},width=0.47\textwidth]{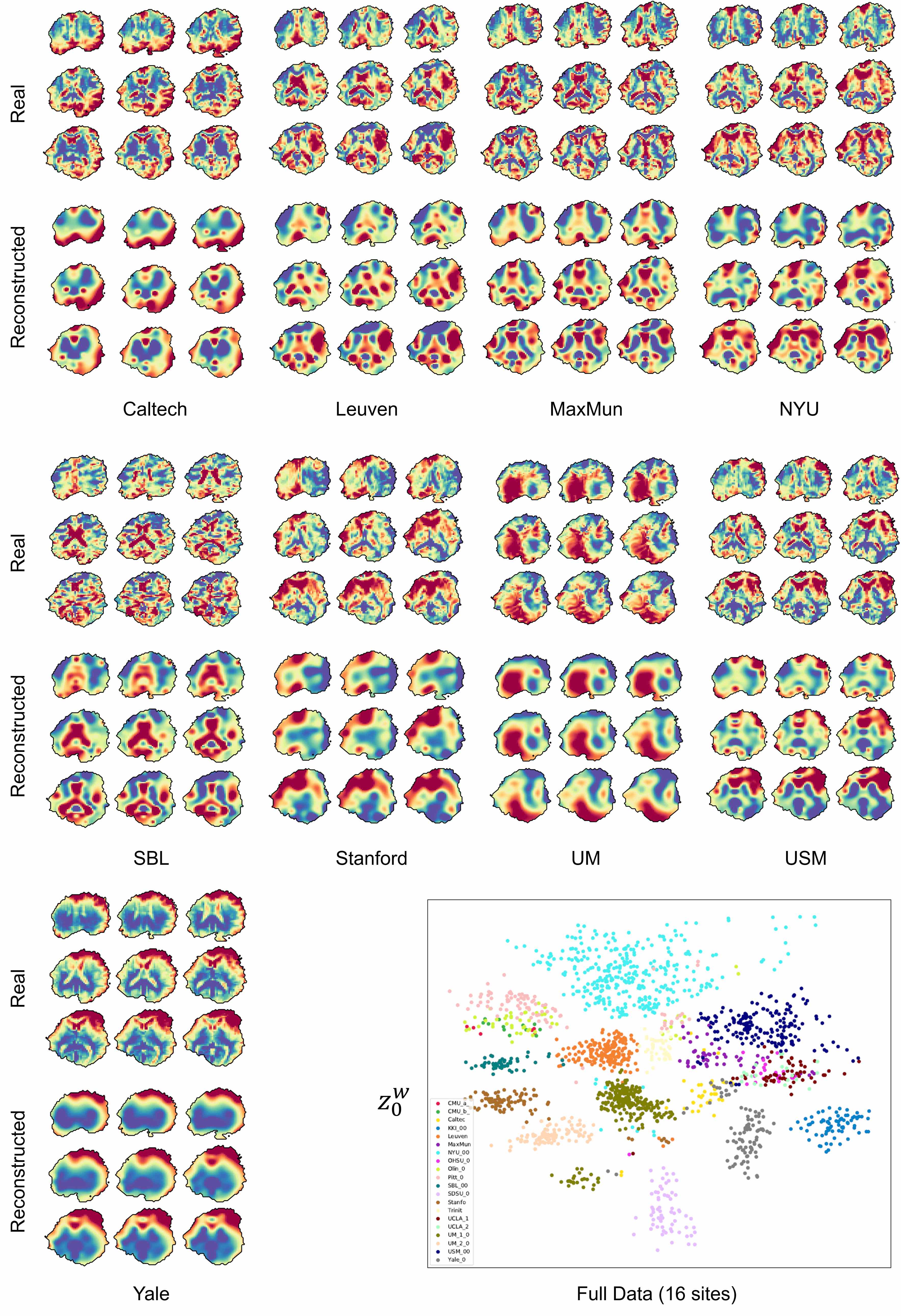}
    \caption{\textbf{Real and reconstructed} brain images from the nine subsets of Autism dataset (Caltex, Leuven, MaxMun, NYU, SBL, Stanford, UM, USM, and Yale sites) showing the smoothing given by sparse factorization. \textbf{Visualizing} $\boldsymbol{z_0^w}$ after training DMSTF on the full autism dataset. DMSTF clustered together temporal latent variables associated with each acquisition site without supervision. As depicted, the variation among different acquisition sites dominates the variation in cognitive state of the brain (ASD group vs. control), hence, a downstream connectivity matrix classification helps better in differentiating ASD group from control in multi-site analysis.}
    \label{fig:picaut}
\end{figure}
}

\newcommand{\picdep}{
\begin{figure*}[t]
    \centering
    \includegraphics[trim={0in 0in 0in 0in},width=1\textwidth]{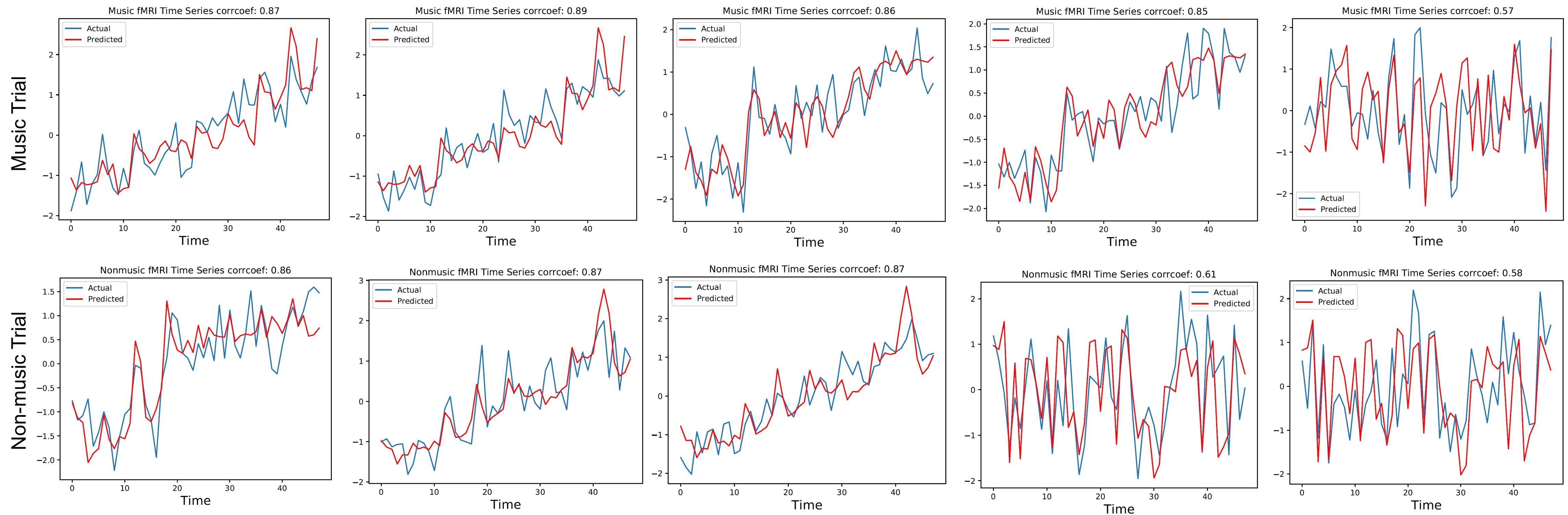}
    \caption{Example fMRI time-series from both musical and non-musical trials (in the test-set of depression dataset) predicted with control variable $u_t=1$ for musical and $u_t=0$ for non-musical trials.}
    \label{fig:picdep}
\end{figure*}
}

\newcommand{\pictraf}{
\begin{figure*}[t]
    \centering
    \includegraphics[trim={0in 0in 0in 0in},width=1\textwidth]{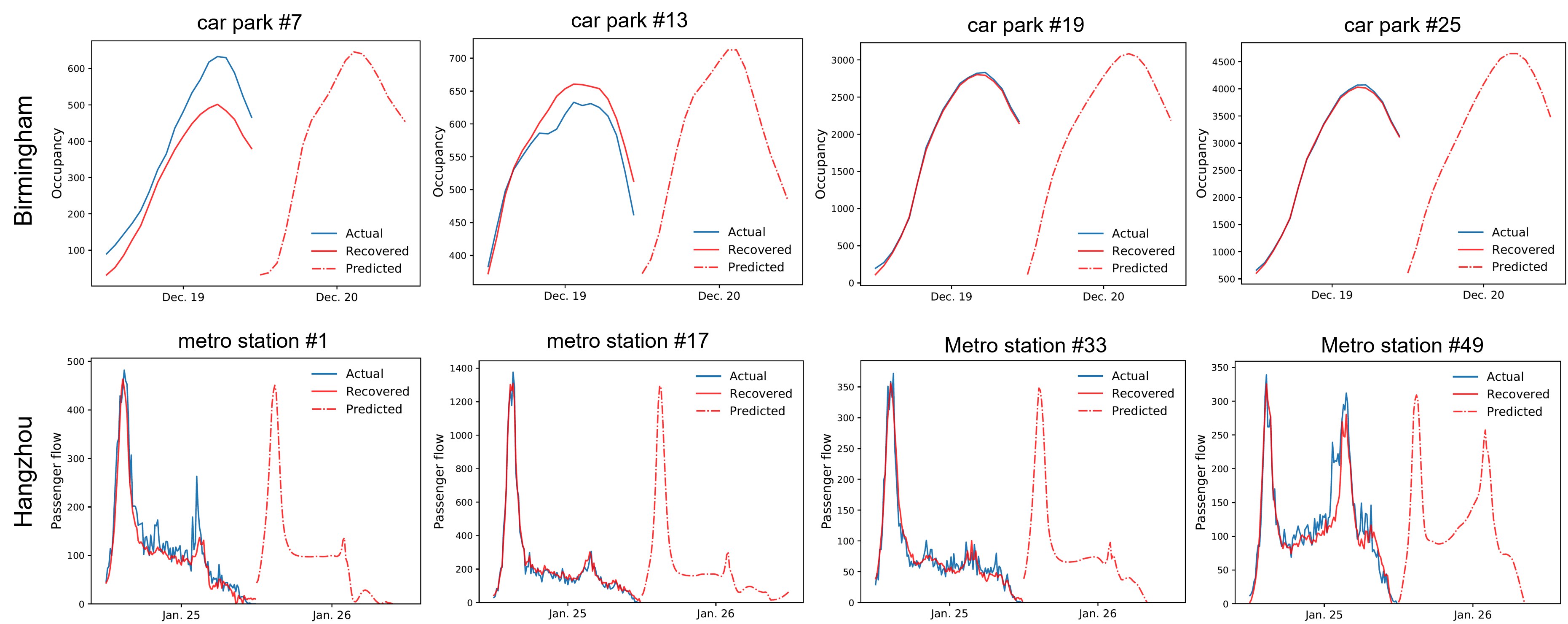}
    \caption{Visualizing next-day (long-term) prediction results for Birmingham and Huangzhou subsets of traffic data for four sample locations. Although DMSTF is well-suited for short-time prediction, next-day forecasts (purely from the trained temporal generative model) show its capability in long-term predictions. Please note that the actual values for the predicted days are not available.}
    \label{fig:pictraf}
\end{figure*}
}

\newcommand{\NetArchneww}{
\begin{table*}[!htbp] 
\centering
\caption{Network architectures for the nonlinear mappings in DMSTF. These fully connected (FC) multilayer perceptron (MLP) networks parameterize the three Gaussian distributions in the generative model: $p_\theta(z^w_t|z^w_{t-1},u_{t-1})$, $p_\theta(w_t|z^w_{t})$ and $p_\theta(H|z^f)$, and the Gaussian variational distribution: $q_\phi(z_t^w|z_{t-1}^w,w_{1:T})$. The second row denotes the inputs to MLPs. The colored numbers indicate intermediate inputs/outputs to/from the corresponding layers. The neural network for $p_\theta(H|z^f)$ can either parameterize the functional form of spatial factors (e.g., $\rho, \gamma$ for Gaussian blobs in fMRI data) or alternatively (ALT.) spatial factors without structural assumptions (e.g., in traffic data).}
{\small
\setlength\tabcolsep{2pt}
\begin{tabular}{|c||c||c||c|}
\toprule
 $\boldsymbol{p_\theta\big(z^w_t \big| [z^w,u]_{t-1}\big)}$ & $\boldsymbol{p_\theta(w_t | z^w_t)}$ & $\boldsymbol{p_\theta(H | z^f)}$ & $\boldsymbol{q(z^w_t|z^w_{t-1},w_{1:T})}$\\
 \toprule
 $\boldsymbol{{[z^w, u]_{t-1}}^{\textcolor{blue}{(1)}} \in\mathbb{R}^{D_z, D_u}}$ & $\boldsymbol{z^w_{t}\in\mathbb{R}^{D_z}}$ & $\boldsymbol{z^f\in\mathbb{R}^{D_z}}$ & $\boldsymbol{{z^w_{t-1}}^{\textcolor{teal}{(1)}},{h_t}\in\mathbb{R}^{D_z,2D_t}}$\\
 \toprule
 \text{FC $(D_z+D_u)\times D_t$ ReLU}  & \text{FC $D_z\times D_e$ ReLU} & \text{FC $D_z\times 2D_z$ ReLU} & \text{${\textcolor{teal}{(1)}}$FC $D_z\times 2D_t^{\textcolor{teal}{(2)}}$}\\
 \text{FC $D_t\times D_z$ Sigmoid} & \text{FC $D_e\times 2D_e$ ReLU} & \text{FC $2D_z\times 4D_z$ ReLU$^{\textcolor{red}{(1)}}$} & \text{$\frac{h_t+\textcolor{teal}{{(2)}}}{2}$FC $2D_t\times D_z$}\\
 $\boldsymbol{g\in \mathbb{R}^{D_z}}$ & \text{FC $2D_e\times 2K$} & \text{FC $4D_z\times 6K$} & $\boldsymbol{\mu_{z^w_t} \in \mathbb{R}^{D_z}}$\\ 
 \text{$\boldsymbol{\textcolor{blue}{(1)}}$FC $(D_z+D_u)\times D_t$ ReLU} & $\boldsymbol{(\mu,\log\sigma)_{w_t}\in\mathbb{R}^{K, K}}$ & $\boldsymbol{\mu_{\rho,\gamma}\in\mathbb{R}^{3K, 3K}}$ & \text{$\frac{h_t+\textcolor{teal}{{(2)}}}{2}$FC $2D_t\times D_z$}\\
 \text{FC $D_t\times D_z^{\textcolor{blue}{(2)}}$} & & \text{${\textcolor{red}{(1)}}$FC $4D_z\times K^{\textcolor{red}{(2)}}$} & $\boldsymbol{\log\sigma_{z^w_t} \in \mathbb{R}^{D_z}}$\\
 $\boldsymbol{\mu^{\text{Nonlinear}}_{z^w_t}\in \mathbb{R}^{D_z}}$& & $\boldsymbol{\log\sigma_{\rho}\in \mathbb{R}^{K}}$ &\\
 \text{$\boldsymbol{\textcolor{blue}{(1)}}$ FC $(D_z+D_u)\times D_z$}& & \text{${\textcolor{red}{(2)}}$ReLU FC $K\times 1$} & \\
 $\boldsymbol{\mu^{\text{Linear}}_{z^w_t}\in \mathbb{R}^{D_z}}$& & $\boldsymbol{\log\sigma_{\gamma}\in \mathbb{R}}$ & \\
 \cline{3-3}\rule{0pt}{2ex}
 \text{${\textcolor{blue}{(2)}}$ ReLU FC $D_z\times D_z$} & & \text{\textcolor{red}{ALT.} ${\textcolor{red}{(1)}}$FC $4D_z\times 2KD$} & \\
  $\boldsymbol{\log\sigma_{z^w_t}\in \mathbb{R}^{D_z}}$ & & $\boldsymbol{(\mu,\log\sigma)_H\in\mathbb{R}^{KD,KD}}$ &\\
 \bottomrule
 \end{tabular}
}
 \label{tbl:NetArchnew}
\end{table*}
}

\newcommand{\nll}{
\begin{table*}[!t] 
\centering
\caption{Held-out Log-Likelihood. DMSTF results in models with higher held-out likelihood, and therefore better fit comparing to HTFA.}
 \begin{tabular}{ l | c | c | c |} 
 \toprule
 Dataset & HTFA & DMSTF \\
 \midrule
 Autism (Caltech) & $-2.82\times10^6$ & $\boldsymbol{-2.33\times10^6}$
 \\
 \midrule
 Depression & $-6.64\times10^5$ & $\boldsymbol{-5.71\times10^5}$
 \\
 \bottomrule
 \end{tabular}
 \label{tbl:heldout}
\end{table*}
}

\newcommand{\traffictab}{
\begin{table*}[!t]
\caption{Performance comparison of short-time prediction. DMSTF outperforms on the test sets of all datasets, doing significantly better particularly on the Birmingham dataset.}
\centering
{\scriptsize
\begin{tabular}{l|cc|cc|cc|cc}
\toprule
\multirow{2}{*}{\backslashbox{Dataset}{Model}} &  \multicolumn{2}{c|}{DMSTF} & \multicolumn{2}{c|}{BTMF} & \multicolumn{2}{c|}{BayesTRMF}& \multicolumn{2}{c}{TRMF} \\
& \tiny{RMSE}   & \tiny{MAPE(\%)}    & \tiny{RMSE}   & \tiny{MAPE(\%)}   & \tiny{RMSE}   & \tiny{MAPE(\%)}      & \tiny{RMSE}   & \tiny{MAPE(\%)}  \\
\midrule
Birmingham   &  \textbf{102.00} & \textbf{20.24}   & 155.32 & 25.10 &  161.11 & 31.80   & 174.25  & 32.63\\
\midrule
Guangzhou   &  \textbf{4.06} & \textbf{10.19}   & 4.09  & 10.25 &  4.27 & 10.70   & 4.30 & 10.65\\
\midrule
Hangzhou   &  \textbf{34.95}  &  29.87   & 37.29  & 30.04 &  40.87 & 30.17   & 39.99  & \textbf{27.77}\\
\midrule
Seattle   &  \textbf{4.49}  &  \textbf{7.39}   & 4.54  & 7.48 &  4.78 & 7.90   & 4.90  & 7.96\\
\bottomrule
\end{tabular}
}
\label{tbl:perf_comp}

\end{table*}

}

\newcommand{\supp}{
\newcommand{\beginsupplement}{%
        \setcounter{table}{0}
        \setcounter{equation}{0}
        \setcounter{section}{0}
        \renewcommand{\thetable}{S\arabic{table}}%
        \setcounter{figure}{0}
        \renewcommand{\thefigure}{S\arabic{figure}}%
}
\renewcommand\thesection{\Alph{section}}
\beginsupplement
\clearpage
\section{More Visualizations for Autism, Depression, and Traffic Datasets}
\label{app:supsecb}
We have visualized real and reconstructed brain images from the nine subsets of autism dataset (Caltex, Leuven, MaxMun, NYU, SBL, Stanford, UM, USM, and Yale sites) along with $z_0^w$ after training DMSTF on the full autism dataset in \figref{picaut}. DMSTF clustered together temporal latent variables associated with each acquisition site without supervision in $z^w_0$. As depicted, the variation among different acquisition sites dominates the cognitive differences between ASD group and control, hence, a downstream connectivity matrix classification (using the learned temporal weights, $W$) helps better in differentiating ASD group from control in multi-site analysis.

\noindent In \figref{picdep}, we have visualized example predicted fMRI time-series from both musical and non-musical trials in the test-set of depression dataset using control variable $u_t=1$ for musical and $u_t=0$ for non-musical trials.

\noindent In \figref{pictraf}, we have visualized next-day (long-term) prediction results for Birmingham and Huangzhou subsets of traffic data for four sample locations. These predictions are purely obtained from the trained temporal generative model. Please note that the actual values for the predicted days are not available.
\vfill
\picaut
\vfill
\picdep
\pictraf
}

\begin{document}

\title{Deep Markov Spatio-Temporal Factorization}

\author{Amirreza Farnoosh$^1$, 
        Behnaz Rezaei$^1$, 
        Eli Sennesh$^2$, Zulqarnain Khan$^1$, Jennifer Dy$^1$, Ajay Satpute$^3$, J. Benjamin Hutchinson$^4$, Jan-Willem van de Meent$^2$
        and~Sarah Ostadabbas$^{1*}$
\IEEEcompsocitemizethanks{\IEEEcompsocthanksitem $^1$A. Farnoosh, B. Rezaei, Z. Khan, J. Dy and S. Ostadabbas are with the Department of Electrical and Computer Engineering, Northeastern University, Boston,
MA, USA.\protect\\
$^*$Corresponding author's email: ostadabbas@ece.neu.edu.
\IEEEcompsocthanksitem $^2$E. Sennesh and J. van de Meent are with the Khoury College of Computer Science, Northeastern University, Boston, MA, USA.
\IEEEcompsocthanksitem $^3$A. Satpute is with the Department of Psychology, Northeastern University, Boston,
MA, USA.
\IEEEcompsocthanksitem $^4$J. Hutchinson is with the Department of Psychology, University of Oregon, Eugene, Oregon, USA.}
}

\markboth{}%
{Shell \MakeLowercase{\textit{et al.}}: Bare Advanced Demo of IEEEtran.cls for IEEE Computer Society Journals}

\IEEEtitleabstractindextext{%
\begin{abstract}
We introduce deep Markov spatio-temporal factorization (DMSTF), a generative model for dynamical analysis of spatio-temporal data. Like other factor analysis methods, DMSTF approximates high dimensional data by a product between time dependent weights and spatially dependent factors. These weights and factors are in turn represented in terms of lower dimensional latents inferred using stochastic variational inference. The innovation in DMSTF is that we parameterize weights in terms of a deep Markovian prior extendable with a discrete latent, which is able to characterize nonlinear multimodal temporal dynamics, and perform multidimensional time series forecasting. DMSTF learns a low dimensional spatial latent to generatively parameterize spatial factors or their functional forms in order to accommodate high spatial dimensionality. We parameterize the corresponding variational distribution using a bidirectional recurrent network in the low-level latent representations. This results in a flexible family of hierarchical deep generative factor analysis models that can be extended to perform time series clustering or perform factor analysis in the presence of a control signal. Our experiments, which include simulated and real-world data, demonstrate that DMSTF outperforms related methodologies in terms of predictive performance for unseen data, reveals meaningful clusters in the data, and performs forecasting in a variety of domains with potentially nonlinear temporal transitions. 
\end{abstract}

}

\maketitle

\IEEEdisplaynontitleabstractindextext

%
\IEEEpeerreviewmaketitle


\ifCLASSOPTIONcompsoc
\IEEEraisesectionheading{\section{Introduction}\label{sec:introduction}}
\else
\section{Introduction}
\label{sec:introduction}
\fi

\IEEEPARstart{A}{nalysis} of large-scale spatio-temporal data is relevant to a wide range of applications in biology, marketing, traffic control, climatology, and neuroscience. 
Due to the high dimensionality of these data, methods for spatio-temporal analysis exploit smoothness and high levels of correlations in the data to map high dimensional data onto a lower dimensional representation \cite{bonilla2008multi,melkumyan2011multi,sun2014collaborative,bahadori2014fast,cai2015facets,zhao2015bayesian,yu2016temporal,takeuchi2017autoregressive,chen2019missing}.

Factor analysis is an established statistical method used to describe variability in high dimensional correlated data in terms of potentially lower dimensional unobserved variables called ``factors''. In other words, factor analysis represents data $Y \in \mathbb{R}^{T \times D}$ with $T$ temporal and $D$ spatial dimensions as a product $Y \simeq W^\top F$ between $K\ll D$ temporal factors (i.e. weights) $W \in \mathbb{R}^{K \times T}$ and spatial factors $F \in \mathbb{R}^{K \times D}$.

In this paper, we present a hierarchical probabilistic factor analysis framework for temporal modeling of high dimensional spatio-temporal data. We explicitly model the correlations between multiple data instances $\{Y_n\}_{n=1}^N$ by representing $W_n$ and $F_n$ in terms of some sets of low dimensional latent variables $Z_n$ that are drawn from shared Markovian and Gaussian priors, respectively.
The result is a modeling framework that uncovers common patterns of multimodal temporal variation across instances, reveals major clusters in the data, and is amenable to temporal forecasting and scalable to very high dimensional data.

Our work is related to recent studies that employ temporal smoothness assumptions \cite{chen2005nonnegative} or more explicitly model dynamics \cite{sun2014collaborative,yu2016temporal,takeuchi2017autoregressive,rogers2013multilinear,jing2018high,bahadori2014fast,cai2015facets} for matrix factorization of multidimensional times series. While most of these methods are not essentially probabilistic (providing only point estimates for imputation/prediction tasks), some Bayesian probabilistic matrix factorization methods have been proposed in \cite{salakhutdinov2008bayesian,zhao2015bayesian,chen2019missing}, and linear temporal dynamics on factor latents, $W$, have been adapted to these methods in  \cite{xiong2010temporal,charlin2015dynamic,sun2019bayesian}. However, these methods do not explicitly adopt \emph{a priori} assumptions about functional form of spatial factors, $F$, when available. This makes them intractable for extremely high dimensional spatial data such as neuroimaging data. Moreover, the linear dynamical assumptions in these methods fall short of modeling potentially nonlinear transitions.

To address these challenges, we introduce deep Markov spatio-temporal factorization (DMSTF)\footnote{The code is available at: \href{https://github.com/ostadabbas/Deep-Markov-Spatio-Temporal-Factorization-DMSTF-}{\textcolor{cobalt}{github.com/ostadabbas/DMSTF}}}, a model that learns a deep generative Markovian prior in order to reason about (potentially nonlinear) temporal dynamics forecasting. This prior can be extended to incorporate a discrete variable for multimodal dynamical estimation and time series clustering, or be conditioned on a control signal that modulates these dynamics. 
In contrast to the previous dynamical matrix factorization methods that model linear temporal dynamics directly in the weights matrix, $W$, DMSTF induces nonlinear temporal dynamics in a lower dimensional latent, $Z$, providing a more tractable model as well as additional interpretable visualizations. At the same time, this model employs a low dimensional spatial latent to generatively parameterize the spatial factors, hence accommodating high dimensionality in both spatial and temporal domains. 

We demonstrate the capabilities of DMSTF in our experiments, which evaluate model performance on simulated data, two large-scale fMRI datasets, and four traffic datasets.
fMRI data are particularly challenging due to their extremely high spatial dimensionality ($D\sim 10^5$), making spatial inductive biases critically necessary. This encourages the need for a low-level hierarchical analysis that summarizes spatial factors with fewer parameters. 
Traffic data, on the other hand, have high temporal dimensionality and often suffer from missing data problem. We show DMSTF to be tractable on these forms of data. Our experiments demonstrate that DMSTF uncovers meaningful clusters in the data, and achieves better predictive performance for unseen data relative to existing baselines, both when evaluating the log-likelihood for new instances and when performing short-term forecasting.
We summarize the contributions of this paper as follows: 
\begin{itemize}
    \item DMSTF learns a deep Markovian prior augmentable with a discrete latent that represents high dimensional spatio-temporal data in terms of low dimensional latent variables that can capture nonlinear multimodal temporal dynamics of the data and perform time series forecasting.
    \item DMSTF employs a low dimensional spatial latent to generatively parameterize spatial factor parameters, hence it accommodates high spatial dimensionality by using a convenient functional form that maps from these parameters to the spatial space.
    \item DMSTF learns a mapping from latent variables to temporal weights and spatial factors, which provide an intermediate representation for downstream regression or classification tasks.
    \item DMSTF is able to perform clustering in the low dimensional temporal latent space, which can provide visual insights about the data.
    \item DMSTF introduces a control input that modulates temporal dynamics, and can be used for learning task-specific temporal generative models in fMRI.
    \item In fMRI analysis, DMSTF was able to partially separate patient and control participants or stimulus types into a low dimensional temporal latent in an unsupervised manner, and resulted in models with higher test set likelihood. Additionally, a downstream classification task on the inferred temporal weights proved better than its anatomically-driven counterpart for patient and control group separation.
    In traffic data, DMSTF outperformed state-of-the-art on short-term prediction of test sets in all datasets.
\end{itemize}

The remainder of this paper is structured as follows. \secref{related} presents an overview of related work. \secref{DMSTF} describes the DMSTF model architecture. \secref{training} provides details on model training, inference and complexity. \secref{experiment} describes experiments on simulated data, neuroimaging datasets, and traffic datasets. \secref{conclusion} discusses conclusions and outlines future work.
\graphrnn

\section{Related Works}
\label{sec:related}
Factor analysis has been extensively used for reducing dimensionality in spatio-temporal data. Principal component analysis (PCA) \cite{pearson1901liii} and independent component analysis (ICA) \cite{comon1991blind} are among the most well-known classical factor analysis methods. To accommodate tensor data, and mitigate scalability issues, multilinear versions of PCA and ICA have been proposed in \cite{hopkins2015tensor,richard2014statistical,cichocki2013tensor,vasilescu2005multilinear}. These methods do not naturally handle missing data. They are also permutation invariant along the batch dimension, and therefore cannot capture temporal dynamics \cite{yu2016temporal}. Spatial factors obtained by these methods are also unstructured, and difficult to interpret in many applications \cite{manning2014topographic}. 

In an early attempt to get temporally smooth structures, Chen and Cichocki \cite{chen2005nonnegative} developed a non-negative matrix factorization model, and applied temporal smoothness and spatial decorrelation regularizers to achieve physiologically meaningful components. Since then, several matrix/tensor factorization approaches have been proposed for modeling temporal dynamics in multivariate/multidimensional time series data. Sun et al. \cite{sun2014collaborative} presented a dynamic matrix factorization suited for collaborative filtering setting in recommendation systems using a linear-Gaussian dynamical state space model. Yu et al. \cite{yu2016temporal} proposed to use an autoregressive temporal regularizer in matrix factorization to describe temporal dependencies in multivariate time series. Takeuchi et al. \cite{takeuchi2017autoregressive} added an additional graph Laplacian regularizer to learn spatial autocorrelations, and perform prediction on unknown locations. Rogers et al. \cite{rogers2013multilinear} applied multilinear dynamical systems to the latent core tensor obtained from Tucker decomposition of tensor time series data, similar to Jing et al. \cite{jing2018high}. Bahador et al. \cite{bahadori2014fast} enforced a low rank assumption on coefficient tensor of vector autoregressive models, and used a spatial Laplacian regularization for prediction in spatio-temporal data. Cai et al. \cite{cai2015facets} developed a probabilistic temporal tensor decomposition that models temporal dynamics in latent factor using a multilinear Gaussian distribution with a multilinear transition tensor and additional contextual constraints.

In contrast to the the methods above, which provide point estimates for imputation/prediction tasks, Bayesian probabilistic matrix/tensor factorization methods have been proposed (see \cite{salakhutdinov2008bayesian,zhao2015bayesian,chen2019missing}). In these approaches, latent factors have Gaussian priors, and Markov chain Monte Carlo (MCMC) methods are used for approximate inference in training and imputation. However, these models essentially focus on global matrix/tensor factorization without explicitly modeling the local temporal and spatial dependencies between factors. Hence, linear temporal dynamics on factor latents have been adapted to Bayesian Gaussian tensor factorization in \cite{xiong2010temporal,charlin2015dynamic,sun2019bayesian}.
While some of these methods have been effective in dynamical modeling of multidimensional time series, they do not explicitly adopt any functional form for the spatial domain, which makes them less effective for high dimensional spatial data such as fMRI. Moreover, the linear dynamical assumptions in these methods lack the capacity to characterize complex nonlinear dependencies.

Motivated by recent advances in deep learning, several papers have studied incorporation of neural networks into Gaussian state space models for nonlinear state space modelling \cite{1511.05121,watter2015embed,krishnan2017structured,fraccaro2017disentangled,karl2016deep,becker2019recurrent}. A common practice in these works is to learn a low dimensional temporal generative model, and a mapping to the observation space, i.e., a decoder, followed by an encoding scheme for performing amortized inference. However, the encoding/decoding framework in these models makes them intractable in high dimensional spatio-temporal data and data with missing values. To be more specific, it is impossible to directly feed the high dimensional data for variational estimation, and also computationally intensive to directly map from the latent space to the high dimensional observation space (see the discussion in \secref{paramcount} for more details).

A number of fMRI-specific hierarchical generative models have been proposed in \cite{manning2014topographic,manning2014hierarchical,manning2018probabilistic,1906.08901}, known as topographic factor analysis (TFA) methods, in which spatial factors are parameterized by Gaussian kernels (i.e., topographic factors) in order to enhance their interpretability.
Among these are hierarchical topographic factor analysis (HTFA) \cite{manning2018probabilistic} which is targeted for multi-subject fMRI analysis, and our prior work, neural topographic factor analysis (NTFA) \cite{1906.08901}, which extends HTFA by incorporating deep generative modeling onto the TFA framework. NTFA assumes separate latent embeddings for participants and stimuli and map them into the temporal and spatial latents with neural networks. However, methods in this line of work assume a prior in which temporal weights are conditionally independent as a function of time, which means that these models do not encode temporal dynamics.

The deep generative Markovian prior employed in DMSTF allows temporal reasoning and forecasting, and is able to model potentially nonlinear temporal transitions. In addition, the spatial generative model with the help of a convenient functional form is able to handle high spatial dimensionality. Finally, the proposed learning and inference strategies make DMSTF framework tractable in both very high dimensional data (e.g., fMRI data) and data with missing values (e.g., traffic data).

In the following section, we formulate deep Markov spatio-temporal factorization in a Bayesian approach and formally explain how the proposed modeling framework is able to address the above-mentioned challenges.

\section{Deep Markov Spatio-Temporal Factorization}
\label{sec:DMSTF}
\subsection{Model Structure and Variational Inference} 
DMSTF defines a hierarchical deep generative model for a corpus of $N$ data instances $\{Y_n\}_{n=1}^N$, as:
\begin{align*}
    Y_n &\sim \text{Norm}(W_n^\top F_n, \: \sigma^\textsc{y} I),
    \\ 
    W_n &\sim \text{Norm}(\mu^\textsc{w}_\theta(Z_n), \sigma^\textsc{w}_\theta(Z_n)),
    \\
    F_n &= \Phi(H_n), \quad\quad
    H_n \sim \text{Norm}(\mu^\textsc{f}_\theta(Z_n), \sigma^\textsc{f}_\theta(Z_n)),
    \\
    Z_n &\sim p_\theta(Z).
\end{align*}
where $p_\theta(Z) $ is a deep generative Markovian prior over a low dimensional set of local (instance-level) variables $Z_n$. The temporal weights $W_n$ are sampled from a Gaussian distribution that is parameterized by neural networks $\mu^\textsc{w}_\theta$ and $\sigma^\textsc{w}_\theta$. The spatial factors are defined as a deterministic transformation $\Phi(H_n)$ of a set of factor parameters $H_n$, which are sampled from a distribution that is parameterized by neural networks $\mu^\textsc{f}_\theta$ and $\sigma^\textsc{f}_\theta$. 
All networks have parameters, which we collectively denote by $\theta$. Finally, $\sigma^\textsc{y}$ denotes the observation noise.

\NetArchneww
We train this model using stochastic variational methods \cite{hoffman2013stochastic,ranganath2013adaptive,kingma2013auto,rezende2015variational}. These methods approximate the posterior $p_\theta(W, H, Z|Y)$ using a variational distribution $q_\phi(W, H, Z)$, where $\phi$ denotes parameters of the variational model, by maximizing a lower bound $\mathcal{L}(\theta,\phi) \le \log p_\theta(Y)$ as:
\begin{align*}
    \mathcal{L}(\theta, \phi)
    &=
    \mathbb{E}_{q_\phi(W, H, Z)}
    \left[
    \log 
    \frac{p_\theta(Y, W, H, Z)}
         {q_\phi(W, H, Z)}
    \right]
    \\
    &=
    \log p_\theta(Y)
    - 
    \text{KL}(q_\phi(W, H, Z) \,||\, p_\theta(W, H, Z|Y)).\numberthis\label{eqn:svi}
\end{align*}

By maximizing this bound with respect to the parameters $\theta$, we learn a deep generative model that defines a distribution over datasets $p_\theta(Y)$, which captures correlations between multiple instances $Y_n$ of the training data. By maximizing the bound with respect to the parameters $\phi$, we perform Bayesian inference by approximating the distribution $q_\phi(W, H, Z) \simeq p_\theta(W, H, Z|Y)$ over latent variables for each instance.

\subsection{Model Variants}
We will develop three variants of DMSTF,  which differ in the set of latent variables that they employ. The graphical models for these model variants are shown in \figref{rnn}. 
All three models incorporate a deep generative Markovian prior to represent temporal variation in $W$. To do so, they introduce a set of time-dependent weight embeddings $z^{w}_t$ to define the distribution on each row, $w_t$, of the weight matrix $W$. Moreover, the $K$ rows of the factor matrix are sampled from a shared prior, which is conditioned on a factor embedding $z^{f}$. The models differ in the following ways:
\begin{itemize}
    \item \textbf{\figref{rnn}(a)}: This model simplifies the structure described above by assuming that the factor parameters $H_n$ and embeddings $z_n^f$ do not vary across instances, but are instead shared at the corpus-level.
    \item \textbf{\figref{rnn}(b)}: This model introduces an additional discrete assignment variable $c$ for each instance, which encourages a multimodal distribution in the temporal latent representations, and serves to perform clustering on time series.
    \item \textbf{\figref{rnn}(c)}: This model additionally introduces a sequence of observed control variables, encoded with one-hot vectors, $\{u_0, \ldots u_{T-1}\}$ that condition the distribution $p_\theta(z^w_t|z^w_{t-1}, u_{t-1})$, therefore govern the temporal generative model.
\end{itemize}

We define and parameterize the generative and variational distributions in \secref{gen_dist} and \secref{var_dist}, respectively. Table~\ref{tbl:NetArchnew} summarizes the neural network architectures we employed in the DMSTF process.

\subsection{Parameterization of the Generative Distributions}
\label{sec:gen_dist}
We structure our generative distribution by incorporating conditional independences inferred from the graphical model of DMSTF in \figref{rnn} (c) as follows:
\begin{align*}
    &Y_n\independent Y_{\neg n}|w_{n,1:T},H_n\quad,\quad z^w_{n,t}\independent z^w_{ n,\neg(t,t-1)}|z^w_{n,t-1}\\ &z^w_{n,t}\independent z^w_{ \neg n}|z^w_{n,t-1}\quad,\quad
    z^w_{n,0}\independent z^w_{\neg n,0}|c_{n}\\
    &w_{n,t}\independent w_{n,\neg t}|z^w_{n,t}\quad,\quad w_{n,t}\independent w_{\neg n}|z^w_{n,t}\quad,\quad H_n\independent H_{\neg n}|z^f_n
\end{align*}

Considering these conditional independencies, the joint distribution of observations and latent variables will be (denoting $Z=\{z^w,z^f\}$):
\begin{align*}
     &p_\theta(Y,c,W,H,Z|u) =\\
     &\prod_{n=1}^N p_\theta(Y_n|w_{n,1:T}, H_n) p_\theta(H_n|z^f_n)p_\theta(z^f_n) p_\theta(z^w_{n,0}|c_n) p_\theta(c_n)\\ 
     &\prod_{t=1}^T p_\theta(w_{n,t}|z^w_{n,t}) p_\theta(z^w_{n,t}|z^w_{n,t-1},u_{n,t-1})
     \numberthis\label{eqn:generative}
\end{align*}

We will parse this proposed generative distribution in the following paragraphs, and provide detailed parameterization of each part.

\vspace{0.5em}\noindent\textbf{Markovian Temporal Latent:} We assume a Gaussian distribution for the latent transition probability. Given the latent $z^w_{t-1}$, 
we parameterize the mean and covariance of the diagonal Gaussian distribution $p_\theta(z^w_t | z^w_{t-1})$ using a neural network. In the model in \figref{rnn}(c) this distribution $p_\theta(z^w_t | z^w_{t-1}, u_{t-1})$ is additionally conditioned on a control variable $u_{t-1}$. Concretely, we pass $z^w_{t-1}$ (and $u_{t-1}$ when applicable) to a multilayer perceptron (MLP) for estimating the Gaussian parameters. We combine a linear transformation of $z^w_{t-1}$ with the estimated mean from the neural network to support both linear and nonlinear dynamics: 
\begin{align*}
    \mu_{z^w_t}  =  (1-g) \odot \mathbb{L}_\theta(z^w_{t-1})+ g \odot \mathbb{F}_\theta(z^w_{t-1}, u_{t-1}),
\end{align*}
where $\mathbb{L}_\theta(\cdot)$ is a linear mapping, $\mathbb{F}_\theta(\cdot)$ is the nonlinear mapping of neural network, $\odot$ denotes element-wise multiplication, and $g\in[0, 1]$ is a scaling vector which itself is estimated from $z^w_{t-1}$ using a neural network.

\vspace{0.5em}\noindent\textbf{Clustering Latent:} In the models in \figref{rnn}(b) and \figref{rnn}(c), we assume that each sequence $Y_n$ belongs to a specific state out of $S$ possible states, and is determined by the categorical variable $c_n$ in our temporal generative model.  This is sampled from a categorical distribution $c_n \sim \text{Cat}(\pi)$, where $\pi = [\pi_1, \cdots, \pi_S]$ specifies cluster assignment probabilities. To this end, we assume that the the first temporal latent $z^w_{n,0}$ is distributed according to a Gaussian mixture: 
\begin{align*}
   p_\theta(z^w_{n,0}|c_n=s) = \text{Norm}(\mu_{s}, \Sigma_{s}),
\end{align*}
where the cluster assignment $c_n$ enforces $\mu_s$ and diagonal covariance $\Sigma_s$.

\vspace{0.5em}\noindent\textbf{Temporal \& Spatial Factors:}
As with the transition model, we assume Gaussian distributions for temporal weights, and spatial factors. We parameterize the diagonal Gaussian distributions for temporal weights $p_\theta(w_t|z^w_t)$ and factor parameters $p_\theta(H|z^f)$  with neural networks. $z^f$ itself is sampled from a normal distribution: $z^f \sim \text{Norm}(0, I)$. Introducing $z^f$ as a low dimensional spatial embedding in the model encourages estimation of a multimodal distribution among spatial factors. 

The form of the spatial factor parameters, $H$, depends on the task at hand. In the case of fMRI data, we employed the construction used in TFA methods \cite{manning2014topographic,manning2014hierarchical,manning2018probabilistic,1906.08901}, which represents each spatial factor as a radial basis function with parameters $H_k=\{\rho_k, \gamma_k\}$:
\begin{align}
    F_{kd}(\rho_k,\gamma_k) = \exp{\Big(-\frac{\|\rho_k-r_d\|^2}{\exp{(\gamma_k)}}\Big)},
\end{align}

This parameterization represents each factor as a Gaussian ``blob''. The vector $r_d \in \mathbb{R}^3$ denotes the position of voxel with index $d$. The parameter 
$\rho_k \in \mathbb{R}^3$ denotes the center of the Gaussian kernel, whereas $\gamma_k \in \mathbb{R}$ controls its width. 

In the case of traffic forecasting experiments, we learned spatial factors without any functional form constraints, hence $H$ directly parameterized $\{F_{kd}\}_{k=1,d=1}^{K,D}$ by mean and covariance of a Gaussian distribution (i.e., $\Phi(\cdot)$ is an identity mapping in this case).

\subsection{Parameterization of the Variational Distributions}
\label{sec:var_dist}
We assume fully-factorized (i.e., mean-field) variational distributions on the variables $\{c,z^f,W,H\}$, and a structured variational distribution on $q_\phi(z^w_{1:T}|w_{1:T})$, hence:
\begin{align*}
     q_\phi(c,W,H,Z|Y,u) =&
     \prod_{n=1}^N q_\phi(c_n) q_\phi(z_n^f) q_\phi(H_n)q_\phi(z_{n,0}^w)\\
     &\prod_{t=1}^T
     q_\phi(z^w_{n,t}|z^w_{n,t-1}, w_{n,1:T}) q_\phi(w_{n,t})
     \numberthis\label{eqn:variational}
\end{align*}
We consider these variational distributions to be Gaussian, and introduce trainable variational parameters $\lambda$, as mean and diagonal covariance of a Gaussian, for each data point in our dataset as follows:
\begin{align*}
\Big\{q(z^w_{n,0};\lambda^w_{n,0}), q(w_{n,t};\lambda^w_{n,t}), q(z_n^f;\lambda^f_n), q(H_n;\lambda^H_{n})
\Big\}_{n=1,t=0}^{N,T}
\end{align*}

We use a structured variational distribution for the variables $q_\phi(z^w_{1:T}|w_{1:T})$ in the form of a one-layer bidirectional recurrent neural network (BRNN) with a rectified linear unit (ReLU) activation, which is then combined with $z^w_{t-1}$ through another neural network to form distribution parameters of $z^w_t$:
\begin{align*}
    &z^w_t 
    \sim \text{Norm}(\mu^w_{t}, \Sigma^w_t),\\ 
    &\{\mu^w_t, \Sigma^w_t\} = f_\phi(z^w_{t-1},h_t),\quad
    h_{1:t} 
    = \textbf{BRNN}_\phi(w_{1:T}).
\end{align*}
where $f_\phi$ is a nonlinear mapping parameterized by an MLP, and $\Sigma$ is diagonal. This structure is encouraged by the Markovian property in the generative model. Note that as it is intractable to directly feed a high dimensional data for variational estimation, we propose to work with the lower dimensional representation vectors, $w_{1:T}$, for this purpose.

Although we can define variational parameters for the categorical distributions $q(c_n)$, we approximate it with the posterior $p(c_n|z^w_{n,0})$ to compensate information loss induced by the mean-field approximation:
\begin{align*}
    q(c_n)\simeq p(c_n|z^w_{n,0})=\frac{p(c_n)p(z^w_{n,0}|c_n)}{\sum_{s=1}^Sp(c_n=s)p(z^w_{n,0}|c_n=s)}.
\end{align*}

This approximation has a two-fold advantage: (1) spares the model additional trainable parameters for the variational distribution, and (2) further links the variational parameters of $q_\phi(z_{n,0}^w)$ to the generative parameters of 
$p_\theta(z^w_{n,0})$ and $p_\theta(c)$, hence results in a more robust learning and inference algorithm.

Derivation of the evidence lower bound for variant (c) of DMSTF is detailed in the following subsection.

\subsection{Evidence Lower BOund (ELBO) for DMSTF}

We derive the ELBO by writing down the log-likelihood of observations, and plugging in $p(\cdot)$ and $q(\cdot)$ from \eqnref{generative} and \eqnref{variational} respectively into \eqnref{svi} (we denote continuous latent variables collectively as $\mathcal{Z}=\{W,H,Z\}$ for brevity):
\begin{align*}
    \mathcal{L}(\theta, \phi)&=
    \mathbb{E}_{q_\phi(c,\mathcal{Z})}\left[ \log\frac{p_\theta(Y,c,\mathcal{Z}|u)}{q_\phi(c,\mathcal{Z})}\right]
    \numberthis\label{eqn:elbo}\\
    &=\mathbb{E}_{q(c,\mathcal{Z})}
    \Bigg[\log\prod\limits_{n=1}^N p(Y_n|w_{n,1:T}, H_n)\\
    &\quad\quad\quad\quad\quad\quad\quad\quad\;
    \frac{p(H_n|z^f_n)p(z^f_n)}{ q(H_n)q(z^f_n)}\frac{p(z^w_{n,0}|c_n)p(c_n)}{q(z^w_{n,0})q(c_n)}\\
    &\quad\quad\quad\quad\quad\quad\;\;
    \prod\limits_{t=1}^T\frac{p(w_{n,t}|z^w_{n,t}) p(z^w_{n,t}|z^w_{n,t-1},u_{n,t-1})}{ q(w_{n,t})q(z^w_{n,t}|z^w_{n,t-1},w_{n,1:T})}\Bigg]
\end{align*}
We further expand the logarithm in \eqnref{elbo} into summation by the product rule, and interchange the summation with the expectation as follows:
{\small
\begin{align*}
    \mathcal{L}(\theta, \phi)=
    \mathlarger{\sum_{n=1}^{N}}&
    \mathbb{E}_{q(w_{n,1:T})q(H_n)}\left[\log p(Y_n|w_{n,1:T},H_n)\right]+\\
    &\mathbb{E}_{q(z^f_n)q(H_n)}\left[\log\frac{p(H_n|z^f_n)}{q(H_n)}\right]+\\
    &\mathbb{E}_{q(z^f_n)}\left[\log\frac{p(z^f)}{q(z^f_n)}\right]+
    \mathbb{E}_{q(c_n)}\left[
    \log\frac{p(c)}{q(c_n)}\right]+\\
    &\mathbb{E}_{q(c_n)q(z^w_{n,0})}\left[\log\frac{p(z^w_{n,0}|c_n)}{q(z^w_{n,0})}\right]+\\
    \sum_{t=1}^T&\mathbb{E}_{q(w_{n,1:T},\, z^w_{n,t-1:t})}\left[\log\frac{p(z^w_{n,t}|z^w_{n,t-1},u_{n,t-1})}{q(z^w_{n,t}|z^w_{n,t-1},w_{n,1:T})}\right]+\\
    &\mathbb{E}_{q(z^w_{n,t})q(w_{n,t})}\left[\log\frac{p(w_{n,t}|z^w_{n,t})}{q(w_{n,t})}\right]
    \numberthis\label{eqn:elbo_expand}
\end{align*}
}
Considering that $\text{KL}(q,p) = \mathbb{E}_{q}\left[\log \frac{q}{p}\right]$, we can rewrite each term in \eqnref{elbo_expand} to summarize the ELBO:
\begin{gather*}
\mathcal{L}(\theta, \phi) = 
\sum_{n=1}^N \Big(\boldsymbol{\mathcal{L}_n^\textbf{rec}}+ \boldsymbol{\mathcal{L}_n^\textbf{H}}+\boldsymbol{\mathcal{L}_n^\textbf{C}}+\sum_{t=1}^T\big(\boldsymbol{\mathcal{L}_{t,n}^{z^w}}+\boldsymbol{\mathcal{L}_{t,n}^{W}}\big)\Big),
\end{gather*}
where,
\begin{align*}
&\boldsymbol{\mathcal{L}_n^\textbf{rec}} = \mathbb{E}_{q_\phi(w_{n,1:T},H_n)}\big[\log p_\theta(Y_n|w_{n,1:T},H_n)\big]\\
&\boldsymbol{\mathcal{L}_{n}^{\textbf{H}}}= -\mathbb{E}_{q_\phi(z^f_n)}\big[\text{KL}\big(q_\phi(H_n)||p_\theta(H_n|z^f_n)\big)\big]\\
&\quad\quad\;\;
-\text{KL}\big(q_\phi(z^f_n)||p_\theta(z^f)\big)\\
&\boldsymbol{\mathcal{L}_n^\textbf{C}} = -\text{KL}\big(q_\phi(c_n)||p_\theta(c)\big)\\
&\quad\quad\;\;
-\sum_{c_n} q_\phi(c_n) \text{KL}\big(q_\phi(z^w_{n,0})||p_\theta(z^w_{n,0}|c_n)\big)\\
&\boldsymbol{\mathcal{L}_{n,t}^{z^w}} = -\mathbb{E}_{q_\phi(w_{n,1:T})}\mathbb{E}_{q_\phi(z^w_{n,t-1}|w_{n,1:T})}\\ 
&\quad\quad\quad\;
\Big[\text{KL}\big(q_\phi(z^w_{n,t}|z^w_{n,t-1},w_{n,1:T})\|
p_\theta\big(z^w_{n,t}|z^w_{n,t-1}, u_{n,t-1})\big)\Big]\\
&\boldsymbol{\mathcal{L}_{n,t}^{\textbf{W}}} = -\mathbb{E}_{q_\phi(z^w_{n,t})}\big[\text{KL}\big(q_\phi(w_{n,t})||p_\theta(w_{n,t}|z^w_{n,t})\big)\big].
\label{eqn:termselbo}\numberthis
\end{align*}
We compute the Monte Carlo estimate of the gradient of the ELBO using a reparameterized sample from the variational distribution of continuous latents. For the discrete latent, $c_n$, we compute the expectations over $q_\phi(c_n)$ by summing over all the possibilities in $c_n$, hence no sampling is performed. We can analytically calculate the KL terms of ELBO for both multivariate Gaussian and categorical distributions, which leads to lower variance gradient estimates and faster training as compared to e.g., noisy Monte Carlo estimates often used in the literature.

\section{Training \& Inference Details}
\label{sec:training}
We described the network architectures for the neural networks used in DMSTF in Table~\ref{tbl:NetArchnew}, where $D_z$ is the dimension of $z^w$ and $z^f$, and $D_t$ and $D_e$ are the dimensions of hidden layers for Markovian and temporal latents respectively. For all the experiments in this paper, we assumed $D_z=2$ (except for traffic dataset). We did all the programming in PyTorch v1.3 \cite{paszke2017automatic}, and used the Adam optimizer \cite{kingma2014adam} with learning rate of $1\times 10^{-2}$. We initialized all the parameters randomly except for spatial locations of Gaussian kernels in fMRI data for which we set the initial values to the local extrema in their averaged fMRI data. We clipped spatial locations and scales to the confines of the brain if needed. We used KL annealing \cite{bowman2015generating} to suppress KL divergence terms in early stages of training, since these terms could be quite strong in the beginning, and we do not want them to dominate the log likelihood term (which controls reconstruction) in early stages. We used a linear annealing schedule to increase from $0.01$ to $1$ over the course of $100$ epochs. We learned and tested all of the models on an Intel Core i7 CPU @3.7 GHz with 8 Gigabytes of RAM, which proves tractability of the learning process. Per-epoch training time varied from $30$ milliseconds in small datasets to $6.0$ minutes in larger experiments, and $500$ epochs sufficed for most of the experiments in the paper.
\subsection{Test/Prediction} 
We report test set prediction error for some of the experiments in this paper. After training DMSTF on a train set, we evaluate the performance of the model in short-term prediction of the test set by adopting a rolling prediction scheme as in \cite{chen2019missing,yu2016temporal}. We predict the next time point of the test set, $\hat{Y}_{t+1}$, using the temporal generative model and spatial factors learned on the train set as follows: $\hat{Y}_{t+1}=\hat{w}_{t+1}^\top F$, where $\hat{w}_{t+1}\sim p(\hat{w}_{t+1}|\hat{z}_{t+1})$, and $\hat{z}_{t+1}\sim p(\hat{z}_{t+1}|z_t)$. Then, we run inference for $Y_{t+1}$, the actual observation at $t+1$, to obtain $z_{t+1}$ and $w_{t+1}$, and use them to predict the next time point, $\hat{Y}_{t+2}$, in the same way. We repeat these steps to make predictions in a rolling manner across a test dataset. We keep the generative model and spatial factors fixed during the entire test set prediction. The root-mean-square error (RMSE) we report for short-term prediction error on a test set is related to the expected negative test set (held-out) log likelihood in our case of Gaussian distributions (with an additive/multiplicative constant), therefore it is used for evaluating the predictive performance of the generative models in the experiments.
\synres
\subsection{Parameter Count for DMSTF}
\label{sec:paramcount}
The number of learnable parameters for the variational distribution in DMSTF is dominated by the parameters of $w_{1:T}$, and therefore will be $O($NTK$)$. 
DMSTF has $O($KD$_e)$ parameters for the temporal generative model, $O($KD$_z)$ parameters for the spatial generative model in fMRI experiments where we use functional form assumptions for spatial factors, and $O($KD$_z$D$)$ parameters for the spatial generative model in traffic experiments without any functional form assumptions for the spatial factors. 
Note that the clustering latent, $c$, does not impose additional parameters to the variational distribution, while only adds $O($SD$_z)$ parameters to the temporal generative model.

\vspace{0.5em}\noindent\textbf{Comparison to Related Works:}
While DMSTF introduces extra features and more complex modeling assumptions for fMRI experiments compared to TFA methods of \cite{manning2014topographic,manning2014hierarchical,1906.08901}, i.e., inferring multimodal nonlinear temporal dynamics and temporal clustering, we want to emphasize that it has the same order of parameters as these methods. TFA methods similarly have $O($NTK$)$ parameters as they employ a fully factorized variational distribution. 

We want to highlight that DMSTF is tractable in both very high dimensional data (e.g., fMRI) and data with missing values in contrast to the previous nonlinear state-space models of \cite{1511.05121,watter2015embed,krishnan2017structured,fraccaro2017disentangled,karl2016deep,becker2019recurrent}. 
This follows from the fact that these works employ an \emph{encoder}, i.e., $q_\phi(w_{n,t}|Y_{n,t})$, to estimate datapoint-specific variational parameters, and this \emph{amortized} framework is not applicable to data with missing entries (without prior imputation) nor extendable to very high dimensional data. Specifically for fMRI experiments, using an encoder/decoder structure as in these works, i.e., $q_\phi(w_{n,t}|Y_{n,t})$, $p_\theta(Y_{n,t}|w_{n,t})$, immediately scales both generative and variational parameters to at least $O($KD$)$, where D $\sim 10^5\gg\text{NT}$, hence causes extensive computational burden and more importantly overfitting. Furthermore, these methods do not learn a generative model for spatial factors, i.e., $p_\theta(H_n)$, and as a result are not able to reason about subject-level variabilities in this respect. We overcome these challenges in DMSTF by carefully designing our non-amortized variational inference and imposing functional form assumptions on the spatial factors in a factorization framework. For the same reason, we conditioned the structured variational distribution of $z^w_{0:T}$ on $w_{1:T}$ in the lower dimensional space rather than the high dimensional observation space (imposed by BRNN as $q_\phi(z^w_{t}|z^w_{t-1}, w_{1:T})$). The proposed learning and inference algorithm keep generative parameters for the high dimensional fMRI data in $O\left(\text{K}(\text{D}_z+\text{D}_e)\right)\ll O(\text{KD})$ as D$_z$, D$_e \sim 2$, and variational parameters in $O($NTK$)$, where NT $\sim 10^2-10^5$, yield an observation to parameter ratio of $O(\frac{\text{NTD}}{\text{NTK}})=O(\frac{D}{K})$ for all the experiments, therefore permit an efficient learning process on large-scale high dimensional data.
 
 
\section{Experimental Results}
\label{sec:experiment}
We analysed the performance of DMSTF on two simulated data, two large-scale neuroimaging datasets and four traffic datasets. First, in a toy example, we showed how DMSTF is able to recover the actual parameters of a nonlinear dynamical system, depicted in \figref{synres} (a). Next, using a synthesized fMRI dataset, we verified the performance of DMSTF in capturing the underlying temporal dynamic and recovering the true clusters in the data, visualized in \figref{synres} (b), (d). Finally, we discussed our results on the real datasets.
\subsection{Toy Example}

We generated $N=100$ synthetic spatio-temporal data with $T=15, K=2$ using a nonlinear dynamical model (motivated by \cite{krishnan2017structured}):
\begin{align*}
&z_t \sim \mathcal{N}\Big(\big[\rho z_{t-1}^0 + \tanh(\alpha z_{t-1}^1)\quad \rho z_{t-1}^1+\sin(\beta z_{t-1}^0)\big],1\Big),\\
&w_t\sim \mathcal{N}(0.5 z_t, 0.1)
\end{align*}
where $\rho=0.2, \alpha=0.5,\beta =-0.1$. For spatial factors, $F$, we picked two Gaussian blobs centered at $\pm(7.5,7.5,7.5)$ with scales of $3, 4.5$ respectively in a box of $30\times30\times30$ at origin. And finally generated $\{Y_n=W_n^\top F\}_{n=1}^{N}$ with additive noise. We trained DMSTF with this synthetic dataset, and estimated the parameters of the model given the functional forms of the generative model. As depicted in \figref{synres} (a) DMSTF was able to recover the actual values of the parameters in this nonlinear dynamical system.
\autres
\subsection{Synthetic Data}
We generated synthetic fMRI data using a MATLAB package provided by \cite{manning2014topographic}, which is known to be useful for analysing fMRI models. The synthesized brain image for each trial (time point) is a weighted summation of a number of radial basis functions (spatial factors) randomly located in the brain. The synthesized fMRI data is then convolved with a hemodynamic response function (HRF), and finally we added zero-mean Gaussian noise with a medium-level signal-to-noise ratio.
Here, we considered $30$ activation sources (spatial factors) randomly located in a standard \texttt{MNI-152-3mm} brain template with roughly $270,000$ voxels, and $150$ trials. We randomly split these $30$ activation sources into $3$ groups, each having $10$ of the Gaussian blobs. These three groups of sources are periodically activated in turn (according to some random weights) for 5 trials.
We generated non-overlapping sequences of $T=5$ time points from this synthetic fMRI data.
This resulted in $10$ data points for each activation group ($N=30$). In order to train DMSTF, we set $T=5$, $K=30$, $D_t=2$, $D_e=8$, $S=3$, and $\sigma_0=1\times 10^{-2}$.
As depicted in \figref{synres} (b), our model was able to successfully recover the $3$ clusters of activation that were present in the dataset in a totally unsupervised manner. The inferred dynamical trajectory of each cluster mean in the temporal latent (i.e., $\mu_{z_t}|\mu_{z_{t-1}}, c$) is visualized in the bottom-right of \figref{synres} (b), and appears to be partitioned into three consecutive rotational dynamics. This is consistent with the periodic activations of sources in data clusters which come in tandem. Predictions of the learned generative model for a selected activation source are visualized in \figref{synres} (d) for the next $50$ time points, estimated as follows: $w_t\sim p(w_t|z_t)$, where $z_t\sim p(z_t|z_{t-1})$ for $t=\{151,\dots,200\}$. These predicted samples perfectly follow hemodynamic response function, confirming DMSTF's capacity in capturing the underlying nonlinear HRF by using MLPs in its temporal generative model. 
\subsection{Neuroimaging Datasets}
We evaluated the performance of DMSTF on a large scale resting-state fMRI data, Autism dataset \cite{craddock2013neuro}, and a task fMRI data, Depression dataset  \cite{lepping2016neural}. We assessed the \emph{clustering} feature of DMSTF on both datasets in terms of disease and cognitive state separation tasks, visualized in \figref{autres} and \figref{depres} (a). Further, we learned task-related temporal generative models for the Depression dataset by incorporating control inputs, $u_{1:T}$, and evaluated them in terms of test set prediction, visualized in \figref{depres} (c), (d). Finally, we provided a quantitative comparison with an established Bayesian generative baseline, HTFA \cite{manning2018probabilistic}, in terms of synthesis quality of the generative models on both datasets by computing held-out (test set) log-likelihood in Table~\ref{tbl:heldout}.
\depres
\nll
\subsubsection{Autism Dataset} We used the publicly available preprocessed resting state fMRI (rs-fMRI) data from the Autism Brain Imaging Data Exchange (ABIDE) collected at 16 international imaging sites \cite{craddock2013neuro}.
This dataset includes rs-fMRI imaging from 408 individuals suffering from Autism Spectrum Disorder (ASD), and 476 typical controls. Each scan has $T=145\sim315$ time points at TR $=2$, and $D=271,633$ voxels. We split the signals into sequences of 75 time points.
We take two approaches to evaluate the performance of our model in separating ASD from control: (1) Cluster data directly in the low dimensional latent, $z^w$, using the clustering feature of DMSTF (called \texttt{DMSTF}), (2) Extract functional connectivity matrices, \cite{hull2017resting}, from learned weights, $W$, followed by a 10-fold SVM for classification (called \texttt{DMSTF+SVM}). As a baseline, we performed a 10-fold SVM classification on extracted connectivity matrices from averaged signals of $116$ regions of interest (ROIs) in automatic anatomical labeling (AAL) atlas \cite{kazeminejad2019topological} (called \texttt{ROI+SVM}).  
Several studies have been done on this dataset to differentiate ASD group from control, all of them using supervised methods, and could achieve accuracies up to 69\% (by carefully splitting data to be as homogeneous as possible, and reducing site-related variability) using the signals extracted from anatomically labeled regions in the brain \cite{kazeminejad2019topological,abraham2017deriving,parisot2017spectral,singh2017constrained}. 
We set $T=75$, $K=100$, $D_e = 15$, $D_t=5$, $S=2$, $\sigma_0 = 1\times 10^{-2}$, and trained DMSTF for 200 epochs on the entire dataset (Full), and also datasets from 9 sites (with more balanced datasets) separately: Caltec, Leuven, MaxMun, NYU, SBL, Stanford, UM, USM, Yale.
As shown in \figref{autres} (c) \texttt{DMSTF} and \texttt{DMSTF+SVM} outperformed \texttt{ROI+SVM} in Caltec, MaxMun, SBL, Stanford, and Yale, while \texttt{ROI+SVM} only performed better in NYU dataset. \texttt{DMSTF+SVM} performed slightly better than \texttt{ROI+SVM} on the entire dataset (Please note that \texttt{DMSTF} is a clustering approach, hence, no error bars are provided in \figref{autres} (c)).
Clustering results for Caltec, Maxmun, SBL, and Stanford are shown in \figref{autres} (a) in which ASD, and control seems to be partially separable (see \figref{picaut} in \appref{supsecb} for more visualization results).
\subsubsection{Depression Dataset}
In this dataset \cite{lepping2016neural}, $19$ individuals with major depressive disorder (MDD) and $20$ never-depressed (ND) control participants listened to standardized positive and negative emotional musical and nonmusical stimuli during fMRI scanning. Each participant underwent 3 musical, and 2 nonmusical runs each for 105 time points at TR=$3$ with $D=353,600$ voxels. During each run, each stimulus type (positive, and negative) was presented for $33$ seconds ($\sim 11$ time points) interleaved with instances of neutral tone of the same length.
We discarded instances of neutral tone, and split each run into non-overlapping sequences of $T=6$ time points in agreement with stimuli design (each stimuli block is split into two sequences).
In other words, each run has $4$ sequences associated with ``positive stimuli'', and $4$ with ``negative stimuli'' resulting in a total of $8$ data points for each run.
In the \textbf{first experiment}, we trained DMSTF on the entire musical runs ($N= 39\times 3\times 8 = 936$) by setting $T = 6$, $K = 100$, $D_e=15$, $D_t=5$, $\sigma_0=1\times 10^{-3}$ for $200$ epochs. The results are shown in \figref{depres} (a, Left). We observed that DMSTF fully separated data points associated with each subject into distinct clusters across the low dimensional temporal latent space. In other words, DMSTF was able to re-unite pieces of signals associated with each subject without any supervision. 
More importantly, DMSTF was able to partially separate data points associated with MDD group from control. As seen in  \figref{depres} (a, Left), MDD group data points are fairly populated in the center of temporal latent while control group are dispersed across latent space. However, DMSTF was not able to meaningfully separate ``negative'' and ``positive'' music pieces in latent embedding from a subject-level perspective, since the variation between runs of a subject dominates stimulus-level variation. For this reason, in a \textbf{second experiment}, we focused on $5$ subjects, and their first musical run from both MDD and control group and trained DMSTF respectively. 
Again, as expected, data points from each subject were distinctly clustered in latent space (see middle and right columns in \figref{depres} (a)). Additionally, DMSTF was able to fit two partially separating Gaussians to ``positive'', and ``negative'' stimuli per subject. 
However, since the number of data points for each subject and run is limited it is not clear how significant these clusters are. A dataset with longer runs could possibly answer that.
In a \textbf{third experiment}, we incorporated control inputs $u_t$, and evaluated the predictive performance of DMSTF in presence of a control signal. We trained DMSTF on two musical runs and a nonmusical run from a subject with depression ($8\times3=24$ sequences) using $u_t=1$, and $u_t=0$ respectively (i.e., $u_t$ is encoding musical, nonmusical stimuli).
We predicted the remaining musical and nonmusical runs ($8\times2=16$ sequences) once with $u_t=0$, and another time with $u_t=1$. As reported in \figref{depres} (d), nonmusical sequences are better predictable with $u_t=0$ than $u_t=1$ with \textit{p}-value of $0.011$ (vice versa for musical sequences with \textit{p}-value of $0.036$). Sample predicted fMRI time series from both musical and nonmusical runs are shown in \figref{depres} (c) (see \figref{picdep} in \appref{supsecb} for more visualization results).

\subsubsection{Comparison with HTFA \cite{manning2018probabilistic}}
We further evaluated DMSTF against HTFA, an established probabilistic generative model for multi-subject fMRI analysis, which uses unimodal Gaussian priors for both temporal weights, and spatial factor parameters, in terms of held-out log-likelihood (see Table~\ref{tbl:heldout}).
For autism, we used Caltech site dataset, and split each subject's fMRI time series into two half (each with $T=70$). We trained DMSTF on the first half, and tested on the second half. For depression dataset, we considered $4$ sequences from each subject's run for training, and tested on the remaining $4$ sequences.
To this end, after training DMSTF on each training set, we fixed the parameters of the generative model, and run inference to obtain variational parameters of the test set for temporal latents $z_t^w, w_t$. And finally computed an importance sampling-based estimate of the log-likelihood \cite{krishnan2017structured}.
The results are shown in Table~\ref{tbl:heldout}, which proves that DMSTF results in models with higher likelihood on test set, hence it is a better fit when compared to HTFA.
\trares
\traffictab
\subsection{Traffic Datasets}
We evaluated the predictive performance of DMSTF against three state-of-the-art baselines on test sets of four traffic datasets. First, we give a brief description of each dataset in the following paragraphs, and then describe the experimental results summarized in Table~\ref{tbl:perf_comp}.

\vspace{0.5em}\noindent\textbf{Birmingham Dataset}\footnote{https://archive.ics.uci.edu/ml/datasets/Parking+Birmingham}: This dataset recorded occupancy of 30 car parks in Birmingham, UK, from October 4 to December 19, 2016 (77 days) every half an hour between 8 a.m. and 5 p.m. (18 time intervals per day) with 14.89\% missing values (completely missing on four days, October 20/21 and December 6/7). We organized the dataset into a tensor of $77\times18\times30$.

\vspace{0.5em}\noindent\textbf{Guangzhou Dataset}\footnote{https://doi.org/10.5281/zenodo.1205229}: This dataset recorded traffic speed from 214 road segments in
Guangzhou, China, from August 1 to September 30, 2016 (61 days) with a 10-minute resolution (144 time intervals per day) with 1.29\% missing values. We organized the dataset into a tensor of $61\times144\times214$.

\vspace{0.5em}\noindent\textbf{Hangzhou Dataset}\footnote{https://tianchi.aliyun.com/competition/entrance/231708/}: This dataset recorded incoming passenger flow from 80 metro stations in Hangzhou, China, from January 1 to January 25, 2019 (25 days) with a 10-minute resolution during service hours (108 time intervals per day). We organized the dataset into a tensor of $25\times108\times80$.

\vspace{0.5em}\noindent\textbf{Seattle Dataset}\footnote{https://github.com/zhiyongc/Seattle-Loop-Data}: This dataset recorded traffic speed from 323 loop detectors in Seattle, USA, over the year of 2015 with a 5-minute resolution (288 time intervals per day). We picked the data from January 1 to January 28 (28 days) as in \cite{chen2019missing}, and organized it into a tensor of $28\times288\times323$.
\subsubsection{Prediction Results}
We compared DMSTF (variant (a)) with three state-of-the-art baselines on our short-term prediction task: TRMF \cite{yu2016temporal}, BayesTRMF developed in \cite{chen2019missing}, and BTMF \cite{chen2019missing}. 
We picked the last seven days from the Birmingham dataset, and the last five days from the Guangzhou, Hangzhou, and Seattle datasets for prediction, then trained the models on the rest for each dataset with $K$=$10,30,10,30$ respectively (consistent setup with \cite{chen2019missing}). For DMSTF, we additionally set $\{D_z$, $D_t$, $D_e\}$ = $5$, $\sigma_0$ = $0$ for all datasets, and learned spatial factors without any functional form constraints. We trained DMSTF for $500$ epochs. We report root mean squared error (RMSE), and mean absolute percentage error (MAPE) for all models on the testsets in Table~\ref{tbl:perf_comp}. DMSTF outperformed in short-term prediction of the test sets on all datasets, doing significantly better particularly on the Birmingham dataset. Testset predictions for two sample locations from each dataset are shown in \figref{trares} (see \figref{pictraf} in \appref{supsecb} for a number of long-term prediction visualizations).

\section{Conclusion and Future Work}
\label{sec:conclusion}
We presented deep Markov spatio-temporal factorization, a new probabilistic model for robust factor analysis of high dimensional spatio-temporal data. We employed a chain of low dimensional Markovian latent variables connected by deep neural networks as a state-space embedding for temporal factors in order to model nonlinear dynamics in data, account better for noise and uncertainty, and enable generative prediction.
We also employed a low dimensional spatial embedding to generate a multimodal distribution of spatial factors. We then demonstrated the tractability of DMSTF on fMRI data (with very high spatial dimensionality) by incorporating functional form assumptions, and on traffic data with high temporal dimensionality. DMSTF enables clustering in the low dimensional temporal latent space to reveal structure in data (e.g., cognitive states in fMRI), providing informative visualizations about the data.
We plan to extend DMSTF to accommodate higher order dynamics for long-term prediction tasks. We can readily achieve this in our setting by conditioning temporal latents on a time-lag set, such as conditioning $z^w_t$ on $z^w_{t-1},z^w_{t-2}$ in a second-order Markov chain.


%





\ifCLASSOPTIONcaptionsoff
  \newpage
\fi

\bibliography{ref}
\bibliographystyle{IEEEtran}
\appendices
\supp
\end{document}